%% file: commitmentAIHRI.tex
\def\year{2019}\relax
\documentclass[letterpaper]{article} 
\usepackage{aaai19}  
\usepackage{times}  
\usepackage{helvet} 
\usepackage{courier}  
\usepackage[hyphens]{url}  
\usepackage{graphicx} 
\usepackage{color}
\usepackage{graphicx}  
\title{Commitments in Human-Robot Interaction}
\author{\Large \textbf{V. Fernandez Castro\textsuperscript{{\rm 1}{\rm 2}}, A. Clodic\textsuperscript{\rm 1}, R. Alami\textsuperscript{\rm 1}, E. Pacherie\textsuperscript{\rm 2}}\\ 
\textsuperscript{\rm 1} LAAS-CNRS, Universit\'e de Toulouse, CNRS, Toulouse, France \\
\textsuperscript{\rm 2} Institut Jean Nicod, CNRS, ENS, PSL Research University, Paris, France \\
victor.fernandez-castro@laas.fr 
}
 
\begin{document}

\maketitle

\begin{abstract}
An important tradition in philosophy holds that in order to successfully perform a joint action, the participants must be capable of establishing commitments on joint goals and shared plans. This suggests that social robotics should endow robots with similar competences for commitment management in order to achieve the objective of performing joint tasks in human-robot interactions. In this paper, we examine two philosophical approaches to commitments. These approaches, we argue, emphasize different behavioral and cognitive aspects of commitments that give roboticists a way to give meaning to monitoring and pro-active signaling in joint action with human partners. To show that, we present an example of use-case with guiding robots and we sketch a framework that can be used to explore the type of capacities and behaviors that a robot may need to manage commitments.
\end{abstract}
\section{Introduction}
A central challenge in human-robot interaction is to devise robots capable of interacting with humans to perform so-called joint actions, social interactions where two or more participants coordinate their actions in space and time to bring about a change in the environment (\citeauthor{sebanz_joint_2006} \citeyear{sebanz_joint_2006}). The notion of commitment plays a pivotal role in understanding joint action \cite{cohen_teamwork_1991,bratman_shared_1992,gilbert_shared_2009}. In order to perform a joint action, co-agents need to establish and handle commitment regarding the shared goals and plans involved in the joint action. However, the study of commitments in the context of human-robot collaborative task achievement is quite new.

As a first approximation, a commitment is in place when an agent (the author of the commitment) has given an assurance to an another agent (the recipient of the commitment) that she will act in a certain way, when, as a result, the recipient has formed  an expectation regarding the actions of the author, and when there is mutual knowledge by both parties that this is the case. In order to build our framework of commitments, we discuss two specific philosophical approaches that emphasize different, though complementary, aspects of commitments: (1) \textit{The functional approach} and (2) \textit{The normative approach}. The first emphasizes the function of commitments, how expectations reduce different forms of uncertainty. The second emphasizes how such a function is carried out; namely, by creating obligations towards one's co-agents and their corresponding entitlements to demand that these obligations be satisfied. In order to demonstrate how these approaches can give hints on the implementation of strategies for HRI, we consider a use-case, in which a robot guide takes a visitor to its destination (e.g. Rackham at the Space City Museum \cite{Clodic_Rackham_2006}). Although there is already previous research on commitments in social robotics, this work is a first step towards generating a framework that gives theoretical support to this previous research and at the same time generates new avenues for future investigations.

\section{Two Philosophical Aspects on Commitments}
Establishing commitments requires the participants involved to generate reliable expectations about the content of the commitment, for instance, making a promise that one will perform a particular objective or giving implicit signals that one act in a particular way. But why do we generate such expectations? Why do we need to establish such commitments? 
 
\subsection{The Functional Approach}
Arguably, what distinguishes a joint action (e.g. walking together) from the mere coordination of behavior (two people randomly walking at the same pace) is that the partners are committed to achieve a goal together \cite{gilbert_2006}. Furthermore, such a participatory commitment comes often with other contralateral commitments \cite{roth2004shared} associated to the general goal or the plan necessary to achieve it (e.g. not speeding up, or waiting for your partner if he's left behind). 

The necessity of establishing these different commitments comes from the fact that even assuming that the partners desire the outcome to be the case, different sources of uncertainty could prevent the joint action from being carried out. As \citeauthor{michael_commitments_2014} (\citeyear{michael_commitments_2014}) argue, participants can face three sources of uncertainty during joint action: 

\begin{enumerate}

\item\textbf{Motivational uncertainty}: participants may not know whether or not the partner is motivated to engage in the joint action. Thus, they cannot be sure how convergent their respective interests are or whether or not they shared a goal. 

\item\textbf{Instrumental uncertainty}: participants may not know whether or not they agree about how to proceed; for instance, how the participants must distribute their roles or when and where they must act.

\item\textbf{Common ground uncertainty}: it may happen that the instrumental beliefs and motivations are not mutually manifested. Thus, even if the participants share a goal or they agree about how to proceed, they might not know that this is the case. 
\end{enumerate}

But how do commitments reduce these sources of uncertainty? According to the \textit{functional approach} supported by \citeauthor{michael_commitments_2014} (\citeyear{michael_commitments_2014}), generating commitments reduces uncertainty by stabilizing expectations regarding actions, beliefs and motivations and thus, facilitating the prediction of the other co-actors, which makes the joint action more fluent and efficient. In their view, the key aspects of commitments are related to how the expectations are generated, and thus, how the commitments are established. For instance, how co-agents generate expectations through implicit and explicit communication \cite{austin_how_1962,siposova_communicative_2018} or repetition \cite{gilbert_2006,michael_sense_2016}.

\subsection{The Normative Approach} 

In a recent paper, \citeauthor{Pacherie_manuscript} (\citeyear{Pacherie_manuscript}) have focused on a different aspect of commitments, namely, how they create obligations towards one's co-agents who are entitled to demand that these obligations be satisfied, giving rise to expectations that the agent will act as committed or that, if not, co-agents will demand that she does. This view shares with the functional approach the idea that expectations are the central element of commitments. However, they argue that the expectations generated by commitments are of a special kind: \textit{normative expectations}.

In philosophy of mind, several authors \cite{greenspan_1978,paprzycka_1999,wallace_1998} have emphasized that we can distinguish two types of expectations:
 
\begin{enumerate}
\item\textbf{Descriptive expectations}: expectations whose violation or frustration does not necessarily triggers reactive attitudes. These expectations are tied to predictions. For example, you can expect your friend to have a beer because this is what she always does but if she doesn't, this may surprise you but not bother you. 

\item\textbf{Normative expectations}: expectations whose violation or frustration triggers reactive attitudes: blame \footnote{The notion of blame must be understood in very general terms and without moral burden in here; that is, as a reaction that serves to sanction negatively a conduct, whether this conduct is a violation of a moral norm or not.}, request for justification or sanctions. These types of expectations are connected to the notion of holding someone on a demand. That is, the normal response to its frustration is to impose a negative reaction to the other agent for not acting as expected. Such negative reactions are more emotionally loaded and aimed at regulating the other's behavior. For instance, you can feel entitled to sanction your friend when he frustrates your expectation that he will cede his seat to an older person on the subway.
\end{enumerate}
 
On \textit{the normative view}, commitments serve to create for oneself, or the other agent, normative expectations, so we can ensure that every participant behaves as expected. In this view, the key aspect of commitments is how we display different communicative and behavioral strategies to make explicit our obligations towards our partner and their own obligations towards us; for instance, by blaming, making the other aware of what is expected or apologizing when something goes wrong. 
 
At this point, it is worth noticing that the two approaches are not necessarily incompatible. In fact, to the extent that they emphasize different aspects of how agents handle commitments, they can be regarded as complementary. In our view, managing commitments involves both generating expectations that facilitate the reduction of uncertainty and displaying repairing and regulative strategies when such expectations are frustrated.

\section{Related work}

Several computational models of commitment management have already been proposed. Joint intention theory \cite{cohen_1990} introduces the notion of commitment in relation to the goal of the collaborative task: ``if a team is jointly committed to some goal, then under certain conditions, until the team as a whole is finished, if one of the members comes to believe that the goal is finished but that this is not yet mutually known, she will be left with a persistent goal to make the status of the goal mutually known". Shared Plans theory \cite{grosz_1996} uses individual intentions to establish commitment of collaborators to their joint activity. Interestingly, Shared Plans theory proposes also the notion of agent's commitments to its collaborating partners' abilities, as well as the notion of helpful behavior and contracting action. Other works emphasize grounding as a kind of commitment \cite{clark_using_1996,clark_social_2006}: ``Once we have formulated a message, we must do more than just send it off. We need to assure ourselves that is has been understood as we intended to be." Clark also proposes to consider on one side \textit{basic joint activity or joint activity proper} and on the other side the \textit{coordinating joint actions} based on \textit{communicative acts} \cite{clark_social_2006}. In the same vein, \citeauthor{klein_2005} (\citeyear{klein_2005}) describes "basic compact" as a kind of contract that need to be shared by the cooperative agents. 

Several implementations using commitments have also been proposed \footnote{In the robotic domain, it is the word \textit{``engagement"} and not \textit{``commitment"} which is often used. For the sake of simplicity we will continue using the term ``commitment" in the paper},  for instance, Collagen architecture \cite{rich_2001,sidner_2005} in collaborative conversation. In human-robot interaction, we can find implementations of commitments  in different settings such as in Human-Robot Interaction Operating System \cite{fong_2005}, robot behavior toolkit \cite{huang_2013}, robot tutelage \cite{Hoffman_2007} or linked to supervision, communicative acts and grounding \cite{Clodic_2007}, \cite{Clodic_2008}.

In the framework of Rackham, a robot guide at the Space City museum\footnote{http://www.laas.fr/robots/rackham} from 2002 to 2005 \cite{Clodic_Rackham_2006}, we also already used the notion of commitments. In that experiment, we defined a set of commitments to the task : Rackham commitment to the task (obtained when the robot decided to do the task), Rackham's belief concerning the visitor's belief about Rackham commitment to the task (obtained once the robot told the visitor that it will perform the task), Rackahm's belief concerning the visitor commitment to the task (obtained once the visitor agrees to do the task). However, commitments were mostly used at task setup and not really handled during task execution. 

The model we propose hereafter takes inspiration from this work. It could be seen as complementary to these models, theories and implementation as it helps to focus on the way we could define and manage commitments during the execution of the task and not just on how commitments can be established at the beginning of the task. The Rackham experiment \cite{Clodic_Rackham_2006} allowed us to note the importance of the quality of interaction. We noticed that: ``a continuous interaction all along the mission is fundamental to keep visitor interest. The robot must continuously check the presence of the guided person, show him that it knows where (s)he is (...)". At that time, it was difficult for us to model that since it was not really part of the task by itself and was done in an intuitive way. In one sense, we think that the theoretical framework we propose here will help us to capture and flesh out this intuition.

\input{new_framework.tex}

\subsection{Future Work}
 
In the previous sections, we have proposed that designing social robots with the capacity for managing commitments can facilitate human-robot interaction. Our central contention is that such a capability, understood as necessary for, but independent of, joint action, plays a central role in the reduction of different types of uncertainties and as a way of enabling participants to react to frustrations of expectations that may undermine the task. Drawing on philosophical approaches to commitments, we have proposed that a commitment management mechanism must be composed of different sub-capabilities for signaling, recognizing signals, monitoring actions, reacting to violations, and repairing different expectations on both the human and robot side. In this section, we would like to propose two lines of development that, we believe, could enrich our framework and open future research avenues.

First, an important way to in which our framework could be extended is by considering the capacity for selecting different signaling strategies depending on the partner. Conforming to expectations or commitments strongly depends on a variety of motivational sources including emotions \cite{fehr_2007,sugden_2000,colombo_2014}, pro-social dispositions \cite{godman_why_2013}, need to belong \cite{Pacherie_manuscript} or reputation management \cite{michael_sense_2016}. These types of motivations are important when deciding on and selecting the appropriate signal for a particular expectation.  For instance, a robot should be capable of deciding on the most appropriate way to ask someone to do something (e.g. using a flat or jovial tone of voice) depending on the assumed motivation in the context (e.g. reputation in a context of work vs pro-social disposition in a playful context). Moreover, ideally, taking into account these motivational components would not only help the robot to select the appropriate signal for a particular expectation but also, to identify and adapt to the particular type of signal used by the human. Being capable of distinguishing someone's motivation and readiness to interact (e.g. distinguish a good teammate from a bad teammate) would provide the robot with better means of adapting and reacting to possible violations during the joint action.   
 
Second, another future direction of research for our framework would be to explore more robust and flexible ways of dealing with the reactions to violations of expectations. Following the functional approach to the notion of commitment, we can advance that an important number of frustrations of expectations would depend on one of the different types of uncertainties (motivation, instrumental or common ground). One plausible improvement of our framework would be dedicated to designing robots capable for determining which source of uncertainty causes the violation and being able to select different repair strategies and reactions depending on the source. To give an example, if Robot could determine the reason why the visitor does not follow it, e.g. motivation: the visitor found a friend; instrumental: the visitor found an obstacle—, it could adapt its answer depending on this factor, for instance, waiting for the human, or approaching her and giving assistance.

\section{Conclusion}

Managing commitments is a central aspect of joint action, and thus, it should be taken into consideration in the design of social robots. In order to do that, we have explored how different philosophical approaches to commitment can provide useful suggestions to roboticists for designing more efficient robots capable of engaging in joint actions. We have presented two complementary proposals that emphasize different behavioral and cognitive aspects of commitments that we must concentrate on for the study of joint action for human-robot interaction.

To design capacities for managing commitments, we must give to the robot abilities to signal its expectations on the one hand and to monitor human’s expectations, reactions and signals on the other. We must also give the human different means of understanding the robot's expectations, reactions and signals on the one hand and of conveying his expectations to the robot on the other. These monitoring and signaling abilities should not be considered part of the action in itself but are needed for the interaction. In this sense perhaps, we should design a kind of code to help the management of these aspects on both the robot's and on human side (e.g. a kind of common interface that is understandable by every robot).

Finally, we have advanced two different research avenues that may enrich our framework. First, one may explore how to design more flexible and adaptable signals depending on the assumed source of motivation behind the joint action. Moreover, such a detection of motivation may help the robots to adapt in case of violation. Second, we have speculated that, when there is a violation of expectation, the source of the violation could come from the different uncertainties explored by the functional approach. As a result, a plausible line of investigation may research how to design the ability to determine the source of violation and adapt depending on the source.

\section{ Acknowledgments}

This work was supported by the ANR JointAction4HRI project (jointaction4hri.laas.fr), grant ANR-16-CE33-0017 of the French Agence Nationale de la Recherche.

\bibliography{JointAction4HRI,biblio}
\bibliographystyle{aaai}

\end{document}

%% file: new_framework.tex
\section{Implementing Commitment}

\subsection{Framework}
We have not yet implemented the proposed framework. Here, we are introducing it using an example, a robot guide such as Rackham at the Space City Museum. In that example, once the destination has been chosen, there should be a joint task of ``go to the destination" where Rackham should ``execute trajectory" whereas the visitor should ``follow the robot". Regarding commitments, we propose that managing commitments is a social skill which requires capacities that are independent of the capacities required to perform the particular actions in place (e.g. \textit{execute trajectory} for Rackham and \textit{follow the robot} for the visitor). In our case, that means that commitment management could be modeled as something aggregated to the actions of the task and linked to expectations.

Given the lessons of the functional and the normative approach, we advance that the function of commitment as a reduction of uncertainty and its normative aspect must be two fundamental elements of the framework. Such elements are present in two different moments of the joint action. First, its function as a reduction tool is exhibited in the repertoire of actions dedicated to insuring the mutual understanding that the commitment is in place before and during the interaction. Second, its normative aspect becomes evident when the commitment is frustrated and certain reactions, responses, and repairs became necessary. In what follows, we present how our framework would depict these two moments. 

\subsection{Generating Commitments} 
\begin{figure}[t]
\centering
\includegraphics[width=0.9\columnwidth]{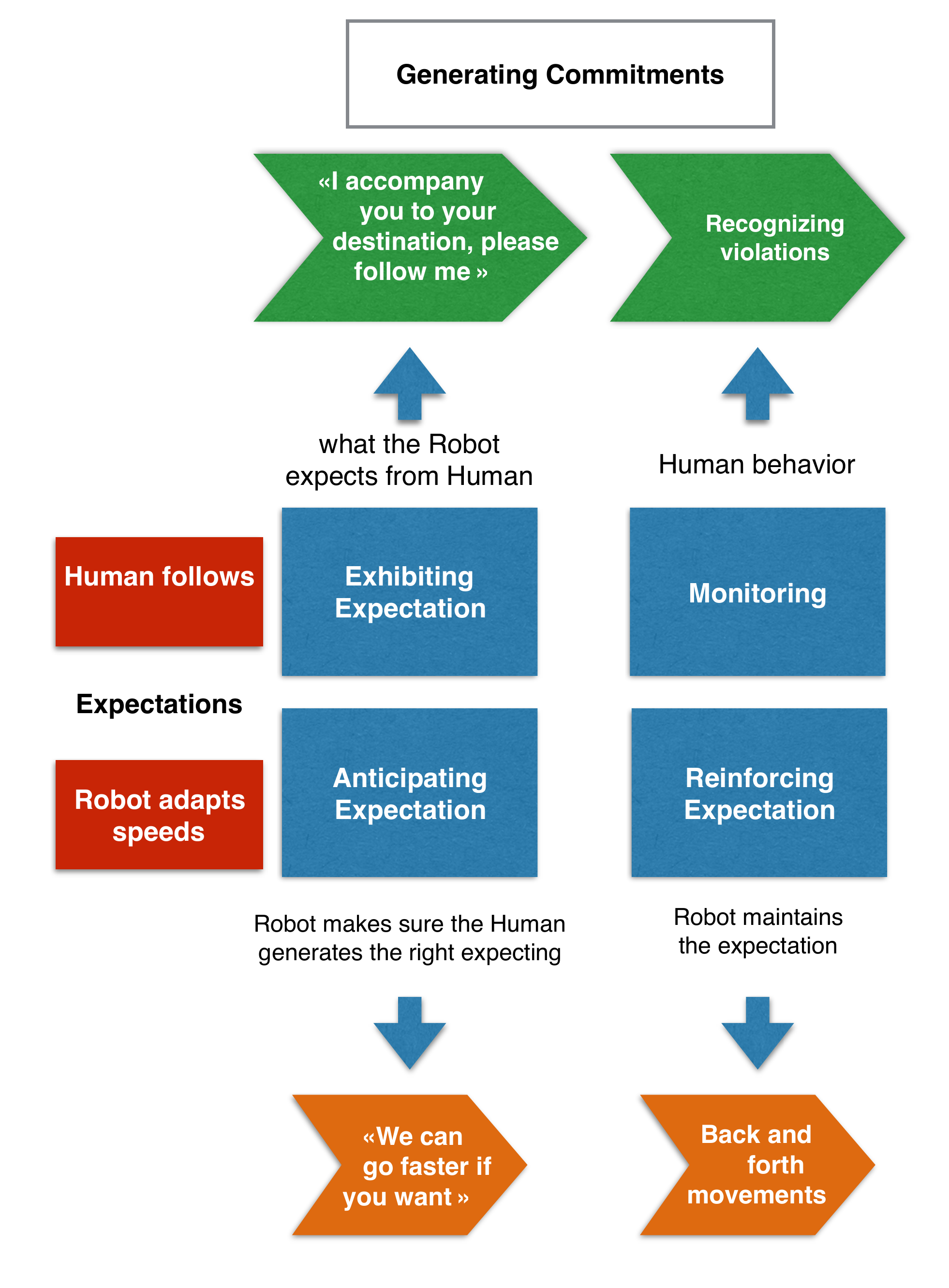} 
\caption{\small{Framework for generating commitments for a guiding task: 
1) When the robot assigns a commitment to the human (the human must follow \textit{(Human follows)}), the following capacities are required: \textbf{Exhibiting expectations} to make the other aware of her obligations; \textbf{Monitoring expectations}: observe and analyse the other agent behavior and detect if she violates expectations. 
2) When the robot assigns a commitment to itself (he must adapt speed \textit{(Robot adapts speed)}), the following capacities are required: \textbf{Anticipating expectations} to make the human aware of the robot’s obligations ; \textbf{Reinforcing expectations}: to make the human aware that the robot that he keeps committed to the task.}} 
\label{figcom1}
\end{figure}

Considering a commitment as a reduction tool implies that ascribing or representing a partner as undertaking a commitment must be tied to a repertoire of actions aiming at insuring the establishment of a mutual understanding of the expectations involved—e.g.  social signals and cues or regulative actions. Such actions can vary depending on whether or not the actor is the agent expecting the other to perform the action (assigning a commitment) or the agent who should act as expected (undertaking the commitment). In our example, Rackham can expect the human to follow him and undertake the commitment of adapting the speed to human pace (which can also be modeled as the human expectation regarding the robot). For a sake of simplicity, from the robot point of view, such difference can be represented in terms of assignations of expectations:
\begin{itemize}	
\item Expectation \textit{(Human follows)}	 
\item Expectation \textit{(Robot adapts speed)} 
\end{itemize}
Given the function of commitments, the robot must be able to actively provide different cues that facilitate human understanding of the expectations in place, but also, he must be able to monitor that the human is behaving as he should (See Figure~\ref{figcom1}). In the guiding example, when Rackham expects the human to follow (Expectation \textit{(Human follows)}), he must possess the following capacities: 
\begin{itemize}
\item \textbf{Exhibiting/Signaling the expectation} Rackham must provide reliable cues regarding what he expects from the visitor in order to make him aware of it. One way of establishing such a signal is through verbal communication, for instance, saying ``I accompany you to your destination, please follow me". 
\item \textbf{Monitoring}: Rackham must be able to recognize the visitor's action and be sure he is behaving as expected, that is, monitoring that the visitor is following.
\end{itemize}

Certainly, the type of reliable signals in place varies with the assignation of expectations. In this case, when the expectation involves the robot itself, he must be sure that the human knows what he is going to do, so the human does not face any type of uncertainty regarding robot’s behavior.  Furthermore, the dynamic nature of joint actions requires the participants to reinforce the pertinent expectations. In our example, thus, assignation of expectations to the human \textit{(Robot adapts speed)} must be tied to the following capacities:
\begin{itemize}
\item \textbf{Anticipating Expectation}: Rackham must make clear that he is capable of adapting to the speed to the human, for instance, using verbal expressions like indirect speech acts (Are you in a rush? We can go faster if you want).
\item \textbf{Reinforcing Expectation}: Rackham must maintain the human expectation that the robot will adapt to his speed. For instance, exhibiting some back and forth movement that makes the human aware that Rackham is still committed to the task.
\end{itemize}

\subsection{Repairing}
Of course, expectations can be frustrated or fulfilled, and this is where the normative aspects of commitments are made explicit. As the normative approach emphasizes, commitments enable participants to hold others on demand and regulate and influence their actions when a particular action is frustrated. This gives to social robotics an important element to improve social interaction by using commitment management as a mechanism for reparation. Given our representation of expectations (Expectation \textit{(Human follows)}/Expectation \textit{(Robot adapts speed)}), we can assume two plausible violations in the context of our example: 
\begin{itemize}
\item The violation of Expectation \textit{(Human follows)}: the visitor does not follow. 
\item The violation of Expectation \textit{(Robot adapts speed)}: the robot is going too slow (or too fast) or the human bump into the robot 
\end{itemize}
Now, considering the first type of violation, the robot must be able to execute two different actions: reaction and repair. The reaction is necessary to make the human understand that the expectation has been frustrated, and thus, that he is responsible for his commitment. Furthermore, such a reaction must be a facilitator for negotiating a re-engagement or repair. In our example, we can implement these two actions as follows:
\begin{itemize}
\item \textbf{Reaction}: when the expectation is frustrated (e.g. the visitor does not follow), Rackham must be able to \textit{react} to the violation in a way that makes the visitor aware that he should be following. For instance, Rackham could stop, look back and adopt a posture of wait-and-see attitude.
\item \textbf{Repair}: Rackham must be equipped with \textit{repairing} strategies that facilitate the re-engagement in the task, for instance, approaching the visitor and asking whether everything is all right. Such a reparation would facilitate the conclusion of the general task, but also, it would make the human aware of his obligations to the robot, reinforcing the motivation for interaction. 
\end{itemize}

Concerning the violation of the robot expectation, the necessary abilities of the robot should differ in several respects. In addition to being able to react and repair in a different way to when it is the human who is at fault, the robot must be able to recognize plausible signals that it is not behaving as expected\footnote{Certainly, we could provide more autonomy to the capacity of the robot to handle commitments by given the capacity of evaluating the situation and recognize that he may not be fulfilling the expectation, and thus, reacting and repairing before the human expresses his discontent.}. In our example, these capacities can be regarded as follows: 
\begin{itemize}
\item \textbf{Recognition of reactions}: when the expectation is frustrated (e.g. Rackham goes too fast or too slow), the robot must be able to notice when the visitor does not follow anymore or, on the contrary, bumps into the robot because it is too slow. Moreover, Rackham must be able to recognize some natural reactions to violation of expectations, like verbal petitions such as ``Hey!" or ``Wait for me!”.
\item \textbf{Reaction}: the recognition that the robot has violated an expectation must trigger an excusing reaction to make the visitor aware that it knows that it has violated the commitment; for instance, apologizing to the visitor.
\item \textbf{Repair}: the reaction must be followed by a repair strategy of the commitment to ensure that the joint action takes place; for instance, slowing down or speeding up according to the human's needs. 
\end{itemize}

\begin{figure}[t]
\centering
\includegraphics[width=0.9\columnwidth]{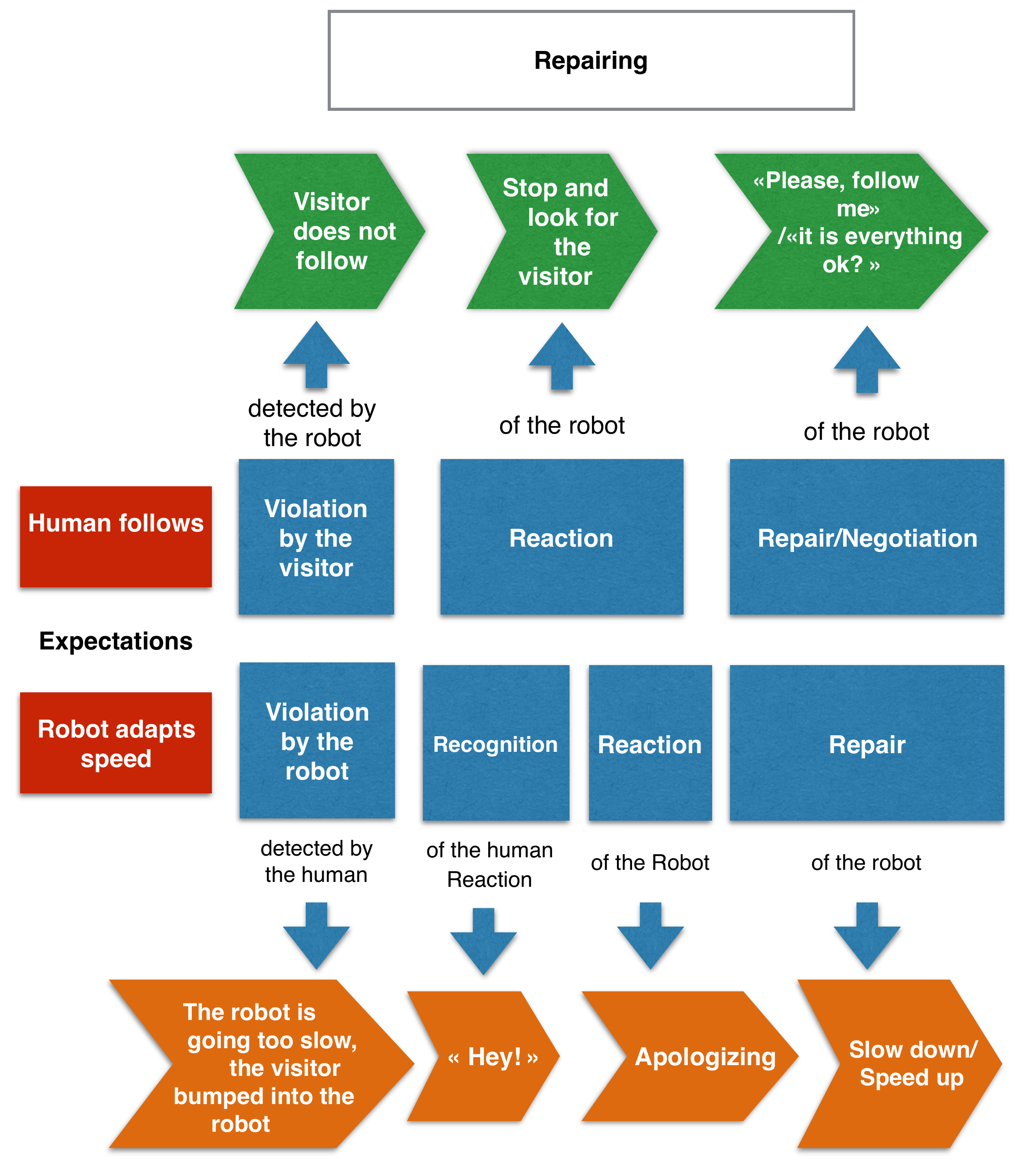} 
\caption{\small{Framework for repairing when a case of violation is detected: 
1) When the robot detects a violation of the human, he must \textbf{React} to  violation  of  expectations  to  make  the  other aware that their obligations are not satisfied; \textbf{Repair} the joint action. 
2) In the case of the robot violation, he must \textbf{Recognize reactions} in order to detect the violation; \textbf{React} to  make the  human aware that he knows that he made a mistake; and \textbf{Repair} the joint action.}} 
\label{figcom2}
\end{figure}

%% file: commitmentAIHRI.bbl
\begin{thebibliography}{}

\bibitem[\protect\citeauthoryear{Austin}{1962}]{austin_how_1962}
Austin, J.
\newblock 1962.
\newblock {\em How to do things with words}.
\newblock Oxford: Clarendon Press.

\bibitem[\protect\citeauthoryear{Bratman}{1992}]{bratman_shared_1992}
Bratman, M.~E.
\newblock 1992.
\newblock Shared {Cooperative} {Activity}.
\newblock {\em Philosophical Review} 101(2):327--341.

\bibitem[\protect\citeauthoryear{Clark}{1996}]{clark_using_1996}
Clark, H.~H.
\newblock 1996.
\newblock {\em Using {Language}}.
\newblock Cambridge: Cambridge University Press.

\bibitem[\protect\citeauthoryear{Clark}{2006}]{clark_social_2006}
Clark, H.~H.
\newblock 2006.
\newblock Social actions, social commitments.
\newblock In Enfield, N., and Levinson, S., eds., {\em Roots of human
  sociality: {Culture}, cognition, and interaction}. New York: Berg.
\newblock  126--150.

\bibitem[\protect\citeauthoryear{Clodic \bgroup et al\mbox.\egroup
  }{2006}]{Clodic_Rackham_2006}
Clodic, A.; Fleury, S.; Alami, R.; Chatila, R.; Bailly, G.; Brethes, L.;
  Cottret, M.; Danes, P.; Dollat, X.; Elisei, F.; and et~al.
\newblock 2006.
\newblock Rackham: An interactive robot-guide.
\newblock In {\em The 15th IEEE International Symposium on Robot and Human
  Interactive Communication, RO-MAN 2006, Hatfield, Herthfordshire, UK, 6-8
  September, 2006},  502–509.
\newblock IEEE.

\bibitem[\protect\citeauthoryear{Clodic \bgroup et al\mbox.\egroup
  }{2007}]{Clodic_2007}
Clodic, A.; Ransan, M.; Alami, R.; and Montreuil, V.
\newblock 2007.
\newblock A management of mutual belief for human-robot interaction.
\newblock In {\em 2007 IEEE International Conference on Systems, Man and
  Cybernetics},  1551–1556.
\newblock IEEE.

\bibitem[\protect\citeauthoryear{Clodic}{2007}]{Clodic_2008}
Clodic, A.
\newblock 2007.
\newblock {\em {Supervision pour un robot interactif: action et interaction
  pour un robot autonome en environnement humain}}.
\newblock Theses, {Universit{\'e} Paul Sabatier - Toulouse III}.

\bibitem[\protect\citeauthoryear{Cohen and Levesque}{1990}]{cohen_1990}
Cohen, P.~R., and Levesque, H.~J.
\newblock 1990.
\newblock Intention is choice with commitment.
\newblock {\em Artificial Intelligence} 42(2):213 -- 261.

\bibitem[\protect\citeauthoryear{Cohen and
  Levesque}{1991}]{cohen_teamwork_1991}
Cohen, P.~R., and Levesque, H.~J.
\newblock 1991.
\newblock Teamwork.
\newblock {\em Nous} 25(4):487--512.

\bibitem[\protect\citeauthoryear{Colombo}{2014}]{colombo_2014}
Colombo, M.
\newblock 2014.
\newblock Caring, the emotions, and social norm compliance.
\newblock {\em Journal of Neuroscience, Psychology, and Economics}.

\bibitem[\protect\citeauthoryear{Fehr and Camerer}{2007}]{fehr_2007}
Fehr, E., and Camerer, C.~F.
\newblock 2007.
\newblock Social neuroeconomics: the neural circuitry of social preferences.
\newblock {\em Trends in Cognitive Sciences} 11(10):419 -- 427.

\bibitem[\protect\citeauthoryear{Fernandez~Castro and
  Pacherie}{2019}]{Pacherie_manuscript}
Fernandez~Castro, V., and Pacherie, E.
\newblock 2019.
\newblock The credibility problem in human robot interaction.
\newblock In {\em 27th Conference of the European Society for Philosophy and
  Psychology}.

\bibitem[\protect\citeauthoryear{Fong \bgroup et al\mbox.\egroup
  }{2005}]{fong_2005}
Fong, T.~W.; Nourbakhsh, I.; Ambrose, R.; Simmons, R.; Schultz, A.; and
  Scholtz, J.
\newblock 2005.
\newblock The peer-to-peer human-robot interaction project.
\newblock In {\em AIAA Space 2005}.

\bibitem[\protect\citeauthoryear{Gilbert}{2006}]{gilbert_2006}
Gilbert, M.
\newblock 2006.
\newblock Rationality in collective action.
\newblock {\em Philosophy of the Social Sciences} 36(1):3–17.

\bibitem[\protect\citeauthoryear{Gilbert}{2009}]{gilbert_shared_2009}
Gilbert, M.
\newblock 2009.
\newblock Shared intention and personal intentions.
\newblock {\em Philosophical Studies} 144(1):167--187.

\bibitem[\protect\citeauthoryear{Godman}{2013}]{godman_why_2013}
Godman, M.
\newblock 2013.
\newblock Why we do things together: {The} social motivation for joint action.
\newblock {\em Philosophical Psychology} 26(4):588--603.

\bibitem[\protect\citeauthoryear{Greenspan}{1978}]{greenspan_1978}
Greenspan, P.~S.
\newblock 1978.
\newblock Behavior control and freedom of action.
\newblock {\em The Philosophical Review} 87(2):225.

\bibitem[\protect\citeauthoryear{Grosz and Kraus}{1996}]{grosz_1996}
Grosz, B.~J., and Kraus, S.
\newblock 1996.
\newblock Collaborative plans for complex group action.
\newblock {\em Artificial Intelligence} 86(2):269 -- 357.

\bibitem[\protect\citeauthoryear{{Hoffman} and {Breazeal}}{2007}]{Hoffman_2007}
{Hoffman}, G., and {Breazeal}, C.
\newblock 2007.
\newblock Effects of anticipatory action on human-robot teamwork: Efficiency,
  fluency, and perception of team.
\newblock In {\em 2007 2nd ACM/IEEE International Conference on Human-Robot
  Interaction (HRI)},  1--8.

\bibitem[\protect\citeauthoryear{Huang and Mutlu}{2013}]{huang_2013}
Huang, C.-M., and Mutlu, B.
\newblock 2013.
\newblock The repertoire of robot behavior: Enabling robots to achieve
  interaction goals through social behavior.
\newblock {\em J. Hum.-Robot Interact.} 2(2):80--102.

\bibitem[\protect\citeauthoryear{Klein \bgroup et al\mbox.\egroup
  }{2005}]{klein_2005}
Klein, G.; Feltovich, P.~J.; Bradshaw, J.~M.; and Woods, D.~D.
\newblock 2005.
\newblock {\em Common Ground and Coordination in Joint Activity}.
\newblock John Wiley and Sons.
\newblock chapter~6,  139--184.

\bibitem[\protect\citeauthoryear{Michael and
  Pacherie}{2014}]{michael_commitments_2014}
Michael, J., and Pacherie, E.
\newblock 2014.
\newblock On {Commitments} and {Other} {Uncertainty} {Reduction} {Tools} in
  {Joint} {Action}.
\newblock {\em Journal of Social Ontology} 1(1):89--120.

\bibitem[\protect\citeauthoryear{Michael, Sebanz, and
  Knoblich}{2016}]{michael_sense_2016}
Michael, J.; Sebanz, N.; and Knoblich, G.
\newblock 2016.
\newblock The {Sense} of {Commitment}: {A} {Minimal} {Approach}.
\newblock {\em Frontiers in Psychology} 6.

\bibitem[\protect\citeauthoryear{Paprzycka}{1999}]{paprzycka_1999}
Paprzycka, K.
\newblock 1999.
\newblock Normative expectations, intentions, and beliefs.
\newblock {\em The Southern Journal of Philosophy} 37(4):629–652.

\bibitem[\protect\citeauthoryear{Rich, Sidner, and Lesh}{2001}]{rich_2001}
Rich, C.; Sidner, C.~L.; and Lesh, N.
\newblock 2001.
\newblock Collagen: Applying collaborative discourse theory to human-computer
  interaction.
\newblock {\em AI Magazine} 22(4):15.

\bibitem[\protect\citeauthoryear{Roth}{2004}]{roth2004shared}
Roth, A.~S.
\newblock 2004.
\newblock Shared agency and contralateral commitments.
\newblock {\em The Philosophical Review} 113(3):359--410.

\bibitem[\protect\citeauthoryear{Sebanz, Bekkering, and
  Knoblich}{2006}]{sebanz_joint_2006}
Sebanz, N.; Bekkering, H.; and Knoblich, G.
\newblock 2006.
\newblock Joint action: bodies and minds moving together.
\newblock {\em Trends in Cognitive Sciences} 10(2):70--76.

\bibitem[\protect\citeauthoryear{Sidner \bgroup et al\mbox.\egroup
  }{2005}]{sidner_2005}
Sidner, C.~L.; Lee, C.; Kidd, C.~D.; Lesh, N.; and Rich, C.
\newblock 2005.
\newblock Explorations in engagement for humans and robots.
\newblock {\em Artificial Intelligence} 166(1):140 -- 164.

\bibitem[\protect\citeauthoryear{Siposova, Tomasello, and
  Carpenter}{2018}]{siposova_communicative_2018}
Siposova, B.; Tomasello, M.; and Carpenter, M.
\newblock 2018.
\newblock Communicative eye contact signals a commitment to cooperate for young
  children.
\newblock {\em Cognition} 179:192--201.

\bibitem[\protect\citeauthoryear{Sugden}{2000}]{sugden_2000}
Sugden, R.
\newblock 2000.
\newblock The motivating power of expectations.
\newblock In Nida-R{\"u}melin, J., and Spohn, W., eds., {\em Rationality,
  Rules, and Structure},  103--129.
\newblock Dordrecht: Springer Netherlands.

\bibitem[\protect\citeauthoryear{Wallace}{1998}]{wallace_1998}
Wallace, R.~J.
\newblock 1998.
\newblock {\em Responsibility and the moral sentiments}.
\newblock Harvard Univ. Press, 2. print edition.

\end{thebibliography}
